\pdfoutput=1

\documentclass[11pt]{article}

\usepackage[]{acl}

\usepackage{times}
\usepackage{latexsym}
\usepackage{times}
\usepackage{latexsym}
\usepackage{graphicx}
\usepackage{amsmath}
\usepackage{amssymb}
\usepackage{threeparttable}
\usepackage{algorithm}
\usepackage{algorithmic}
\usepackage{color,xcolor}
\usepackage{colortbl}
\usepackage{booktabs}

\usepackage[T1]{fontenc}

\usepackage[utf8]{inputenc}
\usepackage{graphicx}
\usepackage{microtype}

\title{Context-aware Difference Distilling for Multi-change Captioning}

\author{Yunbin Tu\textsuperscript{1}, Liang Li\textsuperscript{2,5}\footnotemark[1], Li Su\textsuperscript{1}\footnotemark[1], Zheng-Jun Zha\textsuperscript{3}, \\  \textbf{Chenggang Yan\textsuperscript{4,5},} \textbf{Qingming Huang\textsuperscript{1}} \\ \textsuperscript{1}University of Chinese Academy of Sciences, Beijing, China\\
\textsuperscript{2}Key Lab of Intell. Info. Process., Inst. of Comput. Tech., CAS, Beijing, China\\
\textsuperscript{3}University of Science and Technology of China, Hefei, China\\
\textsuperscript{4}Hangzhou Dianzi University, Hangzhou, China\\ 
\textsuperscript{5}Lishui Institute of Hangzhou Dianzi University, Lishui, China\\
 \texttt{\small{tuyunbin22@mails.ucas.ac.cn},}
 \texttt{\small{liang.li@ict.ac.cn},} \texttt{\small{suli@ucas.ac.cn}}
 }

\begin{document}
\maketitle
\renewcommand{\thefootnote}{\fnsymbol{footnote}}
\footnotetext[1]{Corresponding authors}
\begin{abstract}
Multi-change captioning aims to describe complex and coupled  changes within an image pair in natural language. Compared with single-change captioning, this task requires the model to have higher-level cognition ability to reason an arbitrary number of changes. In this paper, we  propose a novel context-aware difference distilling (CARD) network to capture all genuine changes for yielding sentences. Given an image pair, CARD first decouples   context features that aggregate all similar/dissimilar semantics, termed common/difference context features.  Then, the consistency and independence constraints are designed to guarantee the alignment/discrepancy of common/difference context features. 
Further, the common context features guide  the model to mine locally unchanged features, which are subtracted from the pair to distill locally difference features. Next, the difference context features augment  the locally  difference features to ensure that all changes are distilled. In this way, we obtain an omni-representation of all changes, which is translated into linguistic sentences by a  transformer decoder. Extensive experiments on three public datasets show CARD performs favourably against state-of-the-art methods. The code is available at \url{https://github.com/tuyunbin/CARD}.
\end{abstract}

\section{Introduction}

Change captioning aims to describe differences between a pair of similar images, which enables many important applications, such as automatic report generation about  change conditions of surveillance areas \cite{hoxha2022change} and pathological changes between medical images \cite{liu2021contrastive}.  On the other hand, this task is more challenging than image captioning \cite{yang2023transforming,rotstein2024fusecap,zhao2024cooperative}. This is because machines need to  understand the contents of two images simultaneously, and further reason and caption all  genuine changes between them,  while resisting irrelevant viewpoint/illumination changes. 

\begin{figure}
    \centering
    \includegraphics[width=0.47\textwidth]{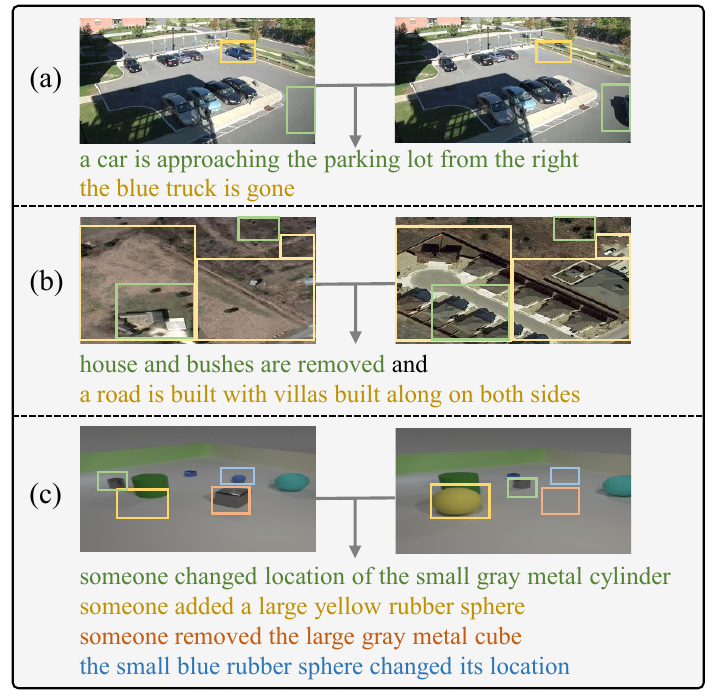}
    \caption{Three examples about multi-change captioning. (a) includes certain object changes; (b) consists of object and background changes; (c) shows both object changes and irrelevant viewpoint change. These changes are shown in colored boxes.}
    \label{fig1}
\end{figure}

Recently, single-change captioning has made remarkable progress \cite{tu2021semantic,hosseinzadeh2021image,yao2022image,yue2023i3n,tu2024smart}. In a dynamic environment, however, the changes are usually the \textit{many-in-one}, where multiple changes exist in an image pair. For instance, there are multiple object/background changes (Figure \ref{fig1} (a) (b)). In other cases, object and viewpoint changes simultaneously appear (Figure \ref{fig1} (c)). In above cases, unchanged objects commonly mingle with changed ones and even appear position misalignment under viewpoint changes. Such distractors pose  a great challenge to identify and caption the genuine  changes. 

There are a few attempts to address multi-change captioning. The pioneer work \cite{jhamtani2018learning} computed pixel differences of two images, which is sensitive to noise. Latest works tried to capture differences at representation space: some of them \cite{hoxha2022change,liu2022remote,liu2023progressive} computed difference features by subtraction, while the others \cite{qiu2021describing,Change2023changes} built the correlations between the patches of two images to model the change features for caption generation.

Despite progress, the above endeavors in multi-change captioning have several limitations. (1) Direct subtraction between two images generalizes poorly to unaligned image pairs under viewpoint changes  (Figure \ref{fig1} (c)). (2) 
Directly correlating  two images fails to sufficiently mine locally unchanged features as multiple objects change, because such features might mingle with the features of changed objects. 
(3) These methods focus on modeling locally difference features, which are useful to catch conspicuous changes. Nevertheless, certain local changes with weak features might be overlooked, \emph{e.g.,} the car occluded by its shadows in Figure \ref{fig1} (a).  These limitations would result in obtaining unreliable difference features for the language decoder. 

We notice that the above methods capture differences between two images only based on local features, while neglecting the use of more comprehensive features. We argue that, to learn locally unchanged/changed features of two images, the model should first  encapsulate their context features of commonality and difference. Such context features aggregate all similar/dissimilar semantics, termed  \textit{common/difference context features}. The former can help correlate and mine locally common features for deducing locally difference features, while  the latter can augment  the locally difference features to ensure all changes are distilled.

In this paper,  we propose a \textbf{C}ontext-\textbf{A}ware Diffe\textbf{R}ence \textbf{D}istilling (CARD) network to learn the robust difference features under multi-change scenes.  Specifically, given the featuers of two images, we first build intra-image interaction to help the model understand each image content of the pair. 
Then, we use CARD  to decouple the common/difference context features from the image  pair. Herein, the common context features of two images summarize joint semantics in between; the difference context feature in each image provides  an independent space to preserve its all changed semantics. Besides, the consistency and independence constraints are designed to enforce the alignment and discrepancy of common and difference context features, respectively. Next, guided by the common context features, CARD models inter-image interaction to mine locally common features, which are removed from the pair to distill locally difference features. Subsequently, CARD augments the locally  difference features via the difference context features, so as to construct an omni-representation of all changes, for generating descriptions by a  transformer decoder.

\textbf{Our key contributions are}: \textbf{(1)} We propose  CARD to first decouple common and difference context features, and then use them to facilitate modeling an omni-representation of all changes for multi-change captioning. \textbf{(2)} The consistency and independence constraints are customized to guarantee the alignment and discrepancy of decoupled common and difference context features. \textbf{(3)}  Extensive experiments show  our method achieves the state-of-the-art results on three public datasets.


\section{Related Work}
Change captioning is an emerging task in the community of multi-modal learning \cite{cong2022ls,cong2023learning,tu20222}. 
In the following, we introduce the relevant works about single-change captioning and multi-change captioning, respectively.  

\textbf{Single-change Captioning} has been widely studied by  most existing methods. The prior work 
\cite{park2019robust} collects a dataset about geometric objects under viewpoint changes. This work computes the difference representation by direct subtraction, which generalizes poorly between two unaligned images. To remedy this limitation, M-VAM \cite{shi2020finding}, VACC \cite{kim2021agnostic} and R$^3$Net \cite{tu-etal-2021-r} match local features  to predict difference features, which has been a classic paradigm. The latest work SCORER+CBR \cite{tu2023self} further improves this paradigm by maximizing cross-view contrastive alignment between two images, so as to learn a more stable difference representation. In addition, these works \cite{hosseinzadeh2021image,kim2021agnostic,tu2023self,yue2024multi} introduce the idea of cross-modal consistency constraint to improve captioning quality. 
Besides improving architecture,  recent works \cite{yao2022image,guo2022clip4idc} propose the strategy of pre-training to fine-tuning for facilitating change location and caption. However, it is seldom that only one change appears in a dynamic environment, so a powerful model should has the capability to describe multiple changes. 
\begin{figure*}[t]
\centering
\includegraphics[width=1\textwidth]{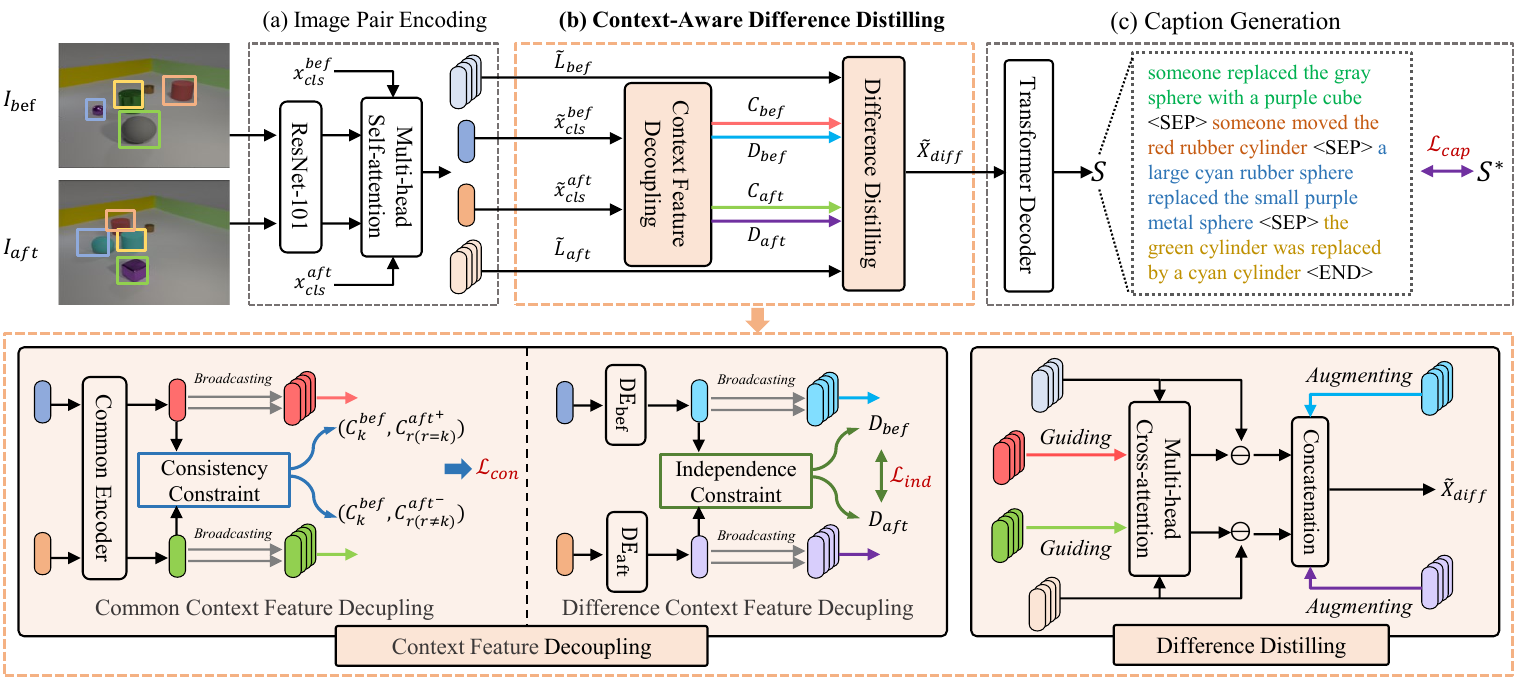} 
  \caption{The overall architecture of our method, which consists of (a) Image Pair Encoding (Sec. \ref{image pair}), (b) \textbf{C}ontext-\textbf{A}ware Diffe\textbf{R}ence \textbf{D}istilling (CARD) (Sec. \ref{clip}), and (c) Caption Generation (Sec. \ref{caption generation}). Herein, CARD is the major component to learn the robust difference features  by context features decoupling and context-aware difference distilling. $S^*$ stands for ground-truth sentences. }
\label{fig2}
\end{figure*}

\textbf{Multi-change Captioning} has been explored by a few attempts. The pioneer work \cite{jhamtani2018learning} proposes to describe multiple changes between the image pairs from the surveillance cameras, where it captures changes at pixel level. Recent works \cite{hoxha2022change,liu2022remote,liu2023progressive,Change2023changes} propose to caption changes between remote sensing images, where they first compute the difference features  and then use them for change detection and description. Nevertheless, the above methods only describe differences between two well-aligned images, while ignoring the cases of unaligned image pairs under varied viewpoints. To this end, Qiu \emph{et al.} \cite{qiu2021describing} collect a dataset to caption multiple  changes under viewpoint changes, where they adopt the classic paradigm of local feature matching to detect changes for caption generation. However,  such a matching paradigm may fail to mine the fine-grained common features and thus learn error-prone difference features. Meanwhile, the current methods focus on modeling local difference features, which risks ignoring certain local changes with weak features.  On the other hand, Qiu \emph{et al.} \cite{qiu2023graph} develop a new synthetic dataset to describe the multiple changes and their orders.  However, recording the order of changed objects is laborious in the real world, making it hard to test such a capability in a real-world scene. 


In short, different from previous methods computing difference features only based on local features,  CARD first decouples common and difference context features from an image pair. The common context features guide the model to fully extract locally unchanged features for computing the features of local differences, while the difference context features  augment the locally difference features to construct an omni-representation of all  changes, for generating accurate sentences.

\section{Methodology}

The overall architecture of our method is shown in Figure \ref{fig2}. Given a pair of images, our method is to generate linguistic sentences that detail all of the changes. Architecture-wise,  
our method contains three components: (a) image pair encoding; (b) context-aware difference distilling; (c) caption generation. We provide an overview of two basic components (a) and (c) in Sec. \ref{image pair} and Sec. \ref{caption generation}, while elaborating our key ingredient (b) in Sec. \ref{clip}.

\subsection{Image Pair Encoding}
\label{image pair}
Formally, given a pair of ``before'' $I_{bef}$ and ``after'' $I_{aft}$ images, we first leverage an off-the-shell encoder (\emph{e.g.,} ResNet-101 \cite{he2016deep}) to extract $n$ local features for each of them, and then introduce a trainable [CLS] feature to represent the global content of each image: $X_{o}=\left\{x_{cls}^{o}, x_{1}, x_{2}, ..., x_{n}\right\}$, where $o \in (bef, aft)$ and $x_{i} \in \mathbb{R}^{d}$. Besides, the trainable position encodings are added into the features of each image to help the model perceive the relative position changes of objects. Next, we exploit a multi-head self-attention layer \cite{vaswani2017attention}  to capture the relationships among the features of each image, which helps the model sufficiently understand the image content of the pair. Through the above manner, we can obtain the relation-embedded features for each image, denoted as $\tilde X_{o}=\left\{\tilde x_{cls}^{o}, \tilde x_{1}, \tilde x_{2}, ..., \tilde x_{n}\right\}$.

\subsection{Context-Aware Difference Distilling}
\label{clip}

\subsubsection{Context Feature Decoupling}

 
 To obtain common and difference context features,  we first devise a common encoder $\mathcal{CE}\left(\cdot ; \theta_{\mathcal{C}}\right)$ and two difference encoders $\mathcal{DE}_o\left(\cdot ; \theta_{o}\right)$, where three encoders are based on the linear projection and  $o \in (bef, aft)$. The common encoder  $\mathcal{CE}\left(\cdot ; \theta_{\mathcal{C}}\right)$ shares the parameters $\theta_{\mathcal{C}}$ between two images, while the difference encoders $\mathcal{DE}_o\left(\cdot ; \theta_{o}\right)$ learn the parameters $\theta_{o}$ for each image. Then, we feed  $\tilde x_{cls}^{o}$ into these encoders to decouple the  common and difference context features, respectively:
\begin{equation}
\begin{aligned}
C_{o}=\mathcal{CE}\left(\tilde x_{cls}^{o}; \theta_{\mathcal{C}}\right), \\
D_{o}=\mathcal{DE}_{o}\left(\tilde x_{cls}^{o}; \theta_{o}\right),
\end{aligned}
\end{equation}
where $C_{bef}$, $C_{aft}$, $D_{bef}$, and $D_{aft} \in \mathbb{R}^{d}$. 

\textbf{Consistency Constraint.} To make two common context features $C_{bef}$ and $C_{aft}$ embedded in a shared space, we tailor the consistency constraint based on contrastive learning. Given a training batch, we sample $B$ pairs of common context features. For the common context feature in the $k$-th ``before'' image  ${C}_{k}^{bef}$,  the common context feature in the $r$-th  ``after'' image ${C}_{r(r=k)}^{aft^+}$ is its positive, while common context features  in the other ``after'' images ${C}_{r(r\neq k)}^{aft^-}$  will be the negatives in this batch. Then, we project these positive/negative pairs into a shared embedding space, normalize them by $L_2$-normalization, and compute their similarity. Next, we introduce the InfoNCE loss \cite{oord2018representation} to optimize their contrastive alignment, \emph{i.e.,} pulling semantically close pairs of  ${C}_{k}^{bef}$ and ${C}_{r(r=k)}^{aft^+}$ together and pushing away non-related pairs:

\begin{equation}
\label{infonce}
\begin{gathered}
\mathcal{L}_{b 2 a}=-\frac{1}{B} \sum_k^B \log \frac{e^{ \left(\text{sim}\left( {C}_{k}^{bef}, {C}_{r(r=k)}^{aft^+}\right) / \tau\right)}}{\sum_r^B e^{ \left(\text{sim}\left({C}_{k}^{bef}, {C}_{r}^{aft}\right) / \tau\right)}}, \\
\mathcal{L}_{a 2 b}=-\frac{1}{B} \sum_k^B \log \frac{e^{ \left(\text{sim}\left( {C}_{k}^{aft}, {C}_{r(r=k)}^{bef^+}\right) / \tau\right)}}{\sum_r^B e^{ \left(\text{sim}\left({C}_{k}^{aft}, {C}_{r}^{bef}\right) / \tau\right)}},\\
\mathcal{L}_{con}=\frac{1}{2}(\mathcal{L}_{b 2 a}+\mathcal{L}_{a 2 b}),
\end{gathered}
\end{equation}
where ``$\text {sim}$'' is the dot-product function to measure the similarity between two context features. $\tau$  is the temperature hyper-parameter. Through this consistency constraint, we enforce the common context features of two images to be projected into a shared semantic space with aligned distributions. 

\textbf{Independence Constraint.} Each decoupled difference context feature can learn the unique characteristics of its corresponding image within the image pair. These unique characteristics represent the semantic differences between the two images, so each difference context feature should be distinct from the other. 
 To this end, we design an independence constraint, which ensures that the two difference context features are projected into separate feature spaces. Here, we opt for the Hilbert-Schmidt Independence Criterion (HSIC)  \cite{song2007supervised}, a proven method for testing feature independence, to design the loss of the independence constraint. A lower HSIC score between the two difference context features indicates a higher independence between them: each difference context feature adequately encapsulates the semantic changes in each image.
 
Concretely, we first project the difference context feature of each image into a separate space, normalize each by $L_2$-normalization, and define the independence (HSIC) constraint  between $D_{bef}$ and $D_{aft}$ as:
\begin{equation}
    \text{HSIC}\left(D_{bef}, D_{aft}\right)=(B-1)^{-2} \operatorname{tr}\left(P K_{bef} P K_{aft}\right),
\end{equation}
where $K_{bef}$, $K_{aft}\in \mathbb{R}^{B\times B}$ are the Gaussian kernel  matrices with $k_{bef, i j}=k_{bef}\left(D_{bef}^i, D_{bef}^j\right)$ and $k_{aft, i j}=k_{aft}\left(D_{aft}^i, D_{aft}^j\right)$. $B$ is batch size. $D_{bef}^i$ refers to the difference context feature in the $i$-th ``befor'' image. $P=\mathbf{I}-\frac{1}{B} e e^T$, where $P\in \mathbb{R}^{B\times B}$, $\mathbf{I}$ is an identity matrix and $e$ is an all-one column vector. If the HSIC score between $D_{bef}$ and $D_{aft}$ is lower, their disparity is more significant.
We define the independence loss as:
\begin{equation}
    \mathcal{L}_{ind}=\text{HSIC}\left(D_{bef}, D_{aft}\right).
\end{equation}

\subsubsection{Difference Distilling}
With the local features of each image $\tilde L_{o}=\left\{\tilde x_{1}, \tilde x_{2}, ..., \tilde x_{n}\right\}$, where $o \in (bef, aft)$, we first broadcast the common context feature  of each image $C_{o} \in \mathbb{R}^{d}$ to $C_{o} \in \mathbb{R}^{n \times d}$. Then, we concatenate it with $\tilde L_{o}$ on the channel dimension to obtain $\tilde X'_{o} \in \mathbb{R}^{n \times 2d}$, where  $C_{o}$ can guide inter-image interaction to mine locally unchanged features. 
Next,  we transform $\tilde X'_{o} \in \mathbb{R}^{n \times 2d}$ to $\tilde X'_{o} \in \mathbb{R}^{n \times d}$ by a non-linear function with ReLU activation. Further, we compute the locally common features on each image by the multi-head cross-attention (MHCA) mechanism \citep{vaswani2017attention}:
\begin{equation}
\begin{gathered}
\tilde{X}_{b e f}^c=\operatorname{MHCA}\left(\tilde{X'}_{b e f}, \tilde{X'}_{a f t}, \tilde{X'}_{a f t}\right), \\
\tilde{X}_{a f t}^c=\operatorname{MHCA}\left(\tilde{X'}_{a f t}, \tilde{X'}_{b e f}, \tilde{X'}_{b e f}\right).
\end{gathered}
\end{equation}
Subsequently, we respectively subtract each $\tilde{X}_{o}^c$ from $\tilde L_{o}$ to compute the locally difference features of  each image. These  locally difference features are further augmented by difference context feature of each image, so as to distill all of the genuine changes in each image:
\begin{equation}
\begin{gathered}
\tilde{X}_{b e f}^d=[\tilde{L}_{b e f} - \tilde{X}_{b e f}^c; D_{bef}], \\
\tilde{X}_{a f t}^d=[\tilde{L}_{aft} - \tilde{X}_{aft}^c; D_{aft}],
\end{gathered}
\end{equation}
where [;] is a concatenation operation. 
Both $\tilde X_{bef}^d$ and $\tilde X_{aft}^d$\ are then concatenated as an omni-representation of all changes between two images, which is implemented by a non-linear transformation with the ReLU  function:
\begin{equation}
\tilde{X}_{d}=\operatorname{ReLU}\left(\left[\tilde{X}_{bef}^{d} ; \tilde{X}_{aft}^{d}\right] W_{c} + b_c \right ).
\end{equation}

\subsection{Caption Generation}
\label{caption generation}
After obtaining the omni-representation  $ \tilde{X}_{d} \in \mathbb{R}^{hw \times d}$, we use a transformer decoder \citep{vaswani2017attention} to decode it into sentences. First, we obtain the embedding features of all $m$ words of these sentences, where each sentence is separated by a special token [SEP]. Then, we use the masked self-attention  to model  relationships among these word features. Next, we model the interaction between the word features and omni-representation by cross-attention, so as to locate the most related difference features during word generation. Subsequently, we feed the selected features into a feed-forward network to obtain the enhanced difference representation, denoted as $\hat H \in \mathbb{R}^{m\times d}$.
Finally, the probability distributions of   words in these sentences are calculated via a single hidden layer:
\begin{equation}
\label{word}
S=\operatorname{Softmax}\left(\hat H W_{s}+{b}_{s}\right),
\end{equation}
where $W_{s}\in \mathbb{R}^{d \times u}$ and $b_{s} \in \mathbb{R}^{u}$ are the learnable parameters. $u$ is the dimension of vocabulary size.

\subsection{Joint Training}

Our method is trained in an end-to-end manner by maximizing the likelihood of the observed  word sequence. Given the ground-truth words $\left(s_{1}^{*}, \ldots, s_{m}^{*}\right)$, we minimize the negative log-likelihood loss:
\begin{equation}
\mathcal L_{cap}(\theta)=-\sum_{t=1}^{m} \log p_\theta \left(s_{t}^{*} \mid s_{<t}^{*}\right),
\end{equation}
where $p_\theta\left(s_{t}^{*} \mid s_{<t}^{*}\right)$ is computed by Eq.~(\ref{word}), and $\theta$ are all the learnable parameters. Our method is also self-supervised by the losses of consistency and independence constraints. Hence, the total loss   is optimized as follows:
\begin{equation}
\label{cross-entropy}
\mathcal L =\mathcal L_{cap} + \lambda_c (\mathcal{L}_{con} + \mathcal{L}_{ind}),
\end{equation}
where $\lambda_{c}$ is a  trade-off parameter to balance the contribution between the caption generator and constraints, which is discussed in the appendix.

\section{Experiments}
\subsection{Datasets}
\textbf{CLEVR-Multi-Change Dataset} \cite{qiu2021describing} is about basic object scene with multiple changes. Since original dataset has not been released, we regenerate this dataset based on the authors' released code. The regenerated dataset has 45,044 valid image pairs/captions with viewpoint changes. 
 Based on the official split, we split it into training, validation, and testing with a ratio of 4:1:1.

\textbf{LEVIR-CC Dataset} \cite{liu2022remote} is about remote sensing scene, which contains 10,077 pairs of bi-temporal images and 50,385 ground-truth captions. 
We use the official split with 6,815 image pairs for training, 1,333 for validation, and 1,929 for testing, respectively.

\textbf{Spot-the-Diff Dataset} \cite{jhamtani2018learning} has 13,192 image pairs from surveillance cameras, and  on an average there are 1.86 ground-truth sentences per image pair.  According to the official split, we split it into training, validation, and testing with a ratio of 8:1:1.



\subsection{Evaluation Metrics}

We follow the existing methods of multi-change captioning to use
the five metrics for evaluating the generated sentences: BLEU-4 (B) \citep{papineni2002bleu}, METEOR (M) \citep{banerjee2005meteor}, ROUGE-L (R) \citep{lin2004rouge}, CIDEr (C) \citep{vedantam2015cider} and SPICE (S) \citep{anderson2016spice}.  We compute all the results by the Microsoft COCO evaluation server \citep{chen2015microsoft}.

\subsection{Implementation Details}
For fair-comparison, we follow previous multi-change captioning methods to use a pre-trained ResNet-101  \citep{he2016deep} to extract the local features of a pair of images, with the dimension of 1024 $\times$ 14 $\times$ 14. We project them into a lower dimension of 512, while the dimension of trainable [CLS] features is also set to 512. The hidden size of the model and word embedding size are set to 512 and 300. Temperature $\tau$ in Eq. (\ref{infonce}) is set to 0.07. 
We train the model to converge with 10K iterations in total. We use Adam optimizer \citep{kingma2014adam} to minimize the negative log-likelihood loss of Eq. (\ref{cross-entropy}). More details are shown in the appendix.

\begin{table}[b]
  \centering
  \caption{Comparison with the SOTA methods on CLEVR-Multi-Change. The main metric
CIDEr on this dataset is highlighted.}
   \small
    \begin{tabular}{c|c|c|c|c|c}
    \toprule
    Method & B     & M     & R    & S & \cellcolor[rgb]{ .859,  .859,  .859}C  \\
    \midrule
    DUDA  & 41.8  & 36.2  & 53.9 & 64.7 & \cellcolor[rgb]{ .859,  .859,  .859}283.5  \\
    M-VAM  & 37.1  & 34.0  & 51.5 & 62.2 & \cellcolor[rgb]{ .859,  .859,  .859}242.9  \\
    MCCFormers-S  & 55.9  & 44.8  & 56.8 & 76.6 & \cellcolor[rgb]{ .859,  .859,  .859}378.6  \\
    MCCFormers-D  & 56.2  & 44.8  & 57.3 & 76.6 & \cellcolor[rgb]{ .859,  .859,  .859}383.2  \\
    VARD-Trans  & 48.1  & 41.8  & 55.5 & 72.1  & \cellcolor[rgb]{ .859,  .859,  .859}344.2 \\
    SCORER+CBR  & 56.4  & 44.9  & 57.1 & 76.7  & \cellcolor[rgb]{ .859,  .859,  .859}388.0  \\
    \midrule
    \textbf{CARD (Ours)} & \textbf{56.7} & \textbf{45.2} & \textbf{57.4} & \textbf{76.9} & \cellcolor[rgb]{ .859,  .859,  .859}\textbf{391.6}  \\
    \bottomrule
    \end{tabular}%
    
  \label{clevr_multi}%
\end{table}%

\subsection{Performance Comparison}

\textbf{Results on CLEVR-Multi-Change.}
We compare CARD with the following SOTA methods: DUDA \cite{park2019robust}, M-VAM \cite{shi2020finding}, MCCFormers-S / MCCFormers-D \cite{qiu2021describing}, VARD-Trans \cite{tu2023viewpoint}, and SCORER+CBR \cite{tu2023self}. On this regenerated dataset, we re-implement the above methods based on their papers and released codes. 

The results are shown in Table \ref{clevr_multi}. Our CARD  performs favourably against these SOTA methods on  all metrics, showing that CARD can better describe multiple changes under viewpoint changes. In addition, our CARD outperforms MCCFormers-D and MCCFormers-S by a large margin, which are classic match-based methods to directly capture inner/inter-patch correlations between two image representations. On the caption-specific metric CIDEr, CARD significantly surpasses both methods, in particular with increases of 2.2\% and 3.4\%.

\begin{table}[t]
  \centering
  \caption{Comparison with the SOTA methods on LEVIR-CC. }
   \small
    \begin{tabular}{c|c|c|c|c}
    \toprule
    Method & B     & M     & R     & \cellcolor[rgb]{ .859,  .859,  .859}C \\
    \midrule
    DUDA   & 57.8  & 37.2  & 71.0  & \cellcolor[rgb]{ .859,  .859,  .859}124.3 \\
    MCCFormers-S    & 56.7  & 36.2  & 69.5  & \cellcolor[rgb]{ .859,  .859,  .859}120.4 \\
    MCCFormers-D  & 56.4  & 37.3  & 70.3  & \cellcolor[rgb]{ .859,  .859,  .859}124.4 \\
    RSICCFormer  & 62.8  & 39.6  & 74.1  & \cellcolor[rgb]{ .859,  .859,  .859}134.1 \\
    PSNet  & 62.1  & 38.8  & 73.6  & \cellcolor[rgb]{ .859,  .859,  .859}132.6 \\
    Prompt-CC (soft) & 62.4  & 38.6  & 73.4  & \cellcolor[rgb]{ .859,  .859,  .859}135.3 \\
    Prompt-CC (hard)  & 63.5  & 38.8  & 73.7  & \cellcolor[rgb]{ .859,  .859,  .859}136.4 \\
    Chg2Cap  & 64.4  & \textbf{40.0}  & \textbf{75.1}  & \cellcolor[rgb]{ .859,  .859,  .859}136.6 \\
    \midrule
    \textbf{CARD (Ours)} & \textbf{65.4} & \textbf{40.0} & 74.6 & \cellcolor[rgb]{ .859,  .859,  .859}\textbf{137.9} \\
    \bottomrule
    \end{tabular}%
  \label{levir}%
\end{table}%

\textbf{Results on LEVIR-CC.}
We compare CARD with the SOTA methods:  DUDA \cite{park2019robust}, MCCFormers-S/D \cite{qiu2021describing}, RSICCFormer \cite{liu2022remote}, PSNet \cite{liu2023progressive}, Prompt-CC (soft/hard) \cite{liu2023Decoupling}, and Chg2Cap \cite{Change2023changes}. 

The experimental results are shown in Table \ref{levir}. Our CARD achieves the best results on all metrics. This indicates that it can detect whether there are semantic changes and what have changed between two remote sensing images.  Besides, we notice that the match-based methods (MCCFormers-D / MCCFormers-S) cannot generalize well in this remote sensing scenario. Our conjecture is that there are usually most changed areas (\emph{e.g.,} Figure \ref{fig1} (b)), so directly matching two images might fail to extract fine-grained unchanged objects and thus capture the difference features with noise. 

\begin{table}[t]
  \centering
   \caption{Comparison with the SOTA methods on Spot-the-Diff. }
  \small
    \begin{tabular}{c|c|c|c|c|c}
    \toprule
    Method & B     & M     & R   & S  & \cellcolor[rgb]{ .859,  .859,  .859}C  \\
    \midrule
    DDLA  & 6.2   & 10.8  & 26.0 & {-} & \cellcolor[rgb]{ .859,  .859,  .859}29.7  \\
    DUDA  & 5.4   & 10.6  & -  & 12.9   & \cellcolor[rgb]{ .859,  .859,  .859}24.8  \\
    MCCFormers-S  & 5.8   & 10.5  & -   & 10.1  & \cellcolor[rgb]{ .859,  .859,  .859}18.2  \\
    MCCFormers-D  & 6.2   & 10.2  & -  & \textbf{17.8}   & \cellcolor[rgb]{ .859,  .859,  .859}28.8  \\
    VARD-Trans  & 4.1   & \textbf{11.4} & 22.2 & 11.5 & \cellcolor[rgb]{ .859,  .859,  .859}13.4  \\
    SCORER+CBR  & 5.1   & 9.3   & 23.0 & 11.9 & \cellcolor[rgb]{ .859,  .859,  .859}20.9  \\
    \midrule
    \textbf{CARD (Ours)} & \textbf{6.6} & 10.8  & \textbf{26.9} & \textbf{17.8} & \cellcolor[rgb]{ .859,  .859,  .859}\textbf{32.4}  \\
    \bottomrule
    \end{tabular}%
  \label{spot}%
\end{table}%

\begin{table*}[htbp]
  \centering
  \small
   \caption{Ablation of common/difference context features (CCF/DCF) on CLEVR-Multi-Change and LEVIR-CC.}
    \begin{tabular}{c|cc|c|c|c|c|c|c|c|c|c}
    \toprule
          &       &       & \multicolumn{5}{c|}{CLEVR-Multi-Change} & \multicolumn{4}{c}{LEVIR-CC} \\
    \midrule
    Ablative Variants & CCF   & DCF   & B     & M     & R  & S   & \cellcolor[rgb]{ .859,  .859,  .859}C      & B     & M     & R     & \cellcolor[rgb]{ .859,  .859,  .859}C \\
    \midrule
    Baseline &   $\times$    &     $\times$  & 54.7  & 43.6  & 56.7  & 75.6 & \cellcolor[rgb]{ .859,  .859,  .859}362.3   & 60.7  & 36.3  & 69.7  & \cellcolor[rgb]{ .859,  .859,  .859}120.0 \\
    Baseline &   $\checkmark$    &     $\times$  & 56.5  & 45.1  & 57.1 & 76.8 & \cellcolor[rgb]{ .859,  .859,  .859}385.8   & 63.5  & 38.5  & 72.3  & \cellcolor[rgb]{ .859,  .859,  .859}130.4 \\
    Baseline &    $\times$   &    $\checkmark$   & 56.5  & 45.0  & 57.1  & \textbf{77.0}  & \cellcolor[rgb]{ .859,  .859,  .859}385.7  & 60.6  & 37.6  & 71.0  & \cellcolor[rgb]{ .859,  .859,  .859}125.9 \\
    Baseline &    $\checkmark$  &    $\checkmark$   & \textbf{56.7} & \textbf{45.2} & \textbf{57.4} & 76.9 & \cellcolor[rgb]{ .859,  .859,  .859}\textbf{391.6}  & \textbf{65.4} & \textbf{40.0} & \textbf{74.6} & \cellcolor[rgb]{ .859,  .859,  .859}\textbf{137.9} \\
    \bottomrule
    \end{tabular}%
   
  \label{module}%
\end{table*}%

\begin{table*}[htbp]
  \centering
  \small
   \caption{Ablation of consistency/independence constraint (CC/IC)  on CLEVR-Multi-Change and LEVIR-CC.}
  \begin{tabular}{c|cc|c|c|c|c|c|c|c|c|c}
    \toprule
          &       &       & \multicolumn{5}{c|}{CLEVR-Multi-Change} & \multicolumn{4}{c}{LEVIR-CC} \\
    \midrule
    Ablative Variants & CC    & IC    & B     & M     & R   & S  & \cellcolor[rgb]{ .859,  .859,  .859}C      & B     & M     & R     & \cellcolor[rgb]{ .859,  .859,  .859}C \\
    \midrule
    CARD  &   $\times$    &     $\times$     & 54.6  & 44.1  & 57.2 & 75.8 & \cellcolor[rgb]{ .859,  .859,  .859}363.7   & 55.9  & 35.6  & 72.3  & \cellcolor[rgb]{ .859,  .859,  .859}132.2 \\
    CARD  &     $\checkmark$   &    $\times$   & 56.2  & 44.8 & 57.1 & 76.8 & \cellcolor[rgb]{ .859,  .859,  .859}384.2   & 56.2  & 35.8  & 72.6  & \cellcolor[rgb]{ .859,  .859,  .859}137.6 \\
    CARD  &    $\times$   &     $\checkmark$   & 56.5  & 45.1 & 57.2 & \textbf{77.0} & \cellcolor[rgb]{ .859,  .859,  .859}389.9   & 60.6  & 37.7  & 72.5  & \cellcolor[rgb]{ .859,  .859,  .859}133.0 \\
    CARD  &     $\checkmark$   &   $\checkmark$     & \textbf{56.7} & \textbf{45.2} & \textbf{57.4} & 76.9 & \cellcolor[rgb]{ .859,  .859,  .859}\textbf{391.6}  & \textbf{65.4} & \textbf{40.0} & \textbf{74.6} & \cellcolor[rgb]{ .859,  .859,  .859}\textbf{137.9} \\
    \bottomrule
    \end{tabular}%
   
  \label{constraint}%
\end{table*}%

\textbf{Results on the Spot-the-Diff Dataset.}
Most previous works \cite{hosseinzadeh2021image,huang2022image,tu2023neighborhood,yue2023i3n} tested the models based on single-change setup, where the models are only required to randomly describe one of the changes. Different from them, we require the model to caption all changes.

Under this setting, the compared SOTA methods are: DDLA \cite{jhamtani2018learning}, DUDA \cite{park2019robust}, MCCFormers-S / MCCFormers-D \cite{qiu2021describing}, VARD-Trans \cite{tu2023viewpoint}, and SCORER+CBR \cite{tu2023self}. The experimental results are shown in Table \ref{spot}. Our method achieves the best results on most metrics, particularly with an increase of 9.1\% on CIDEr. As shown in Figure \ref{fig1} (a), the changed objects are not well captured by surveillance cameras and are even occluded by shadows.  Our method still achieves encouraging performance, which validates its good generalization in surveillance scenes.

\textbf{Performance Analysis.} On the three datasets, our CARD outperforms the existing methods  by a large margin, which shows its good generalization. This superiority  benefits from that decoupled context features facilitate learning difference features. Instead, the compared methods only compute the differences based on the local features, which fails to mine fine-grained common objects  and thus disregards inconspicuous changes. 

\subsection{Ablation Study and Analysis}
To figure out the contribution of each component, we conduct ablation studies on two large-scale datasets: CLEVR-Multi-Change and LEVIR-CC. The image pairs on CLEVR-Multi-Change contain basic geometric objects and  are unaligned due to viewpoint changes, while the pairs on LEVIR-CC are well-aligned and from the real world.

\subsubsection{Ablation Study for Context Features} We study the effectiveness of  decoupled common and difference context features, denoted as CCF and DCF, respectively. The results are shown in Table \ref{module}.  Baseline directly matches two image features to extract the locally common features and difference features for caption generation.

We find that on the both datasets, 1) the model's performance is enhanced when it is augmented by either CCF or DCF; 2) when we augment the model with both context features, the model's performance is significantly boosted, especially the CIDEr score is enhanced from 362.3 to 391.6. These show that 1) CCF guides the model to sufficiently interact and mine locally common features for computing  locally difference features, while DCF augments the locally difference features to ensure all changes are distilled; 2) each kind of context feature not only plays its unique role, but also supplements each other for better reasoning genuine changes. 

\begin{figure}[t]
\centering
\includegraphics[width=0.47\textwidth]{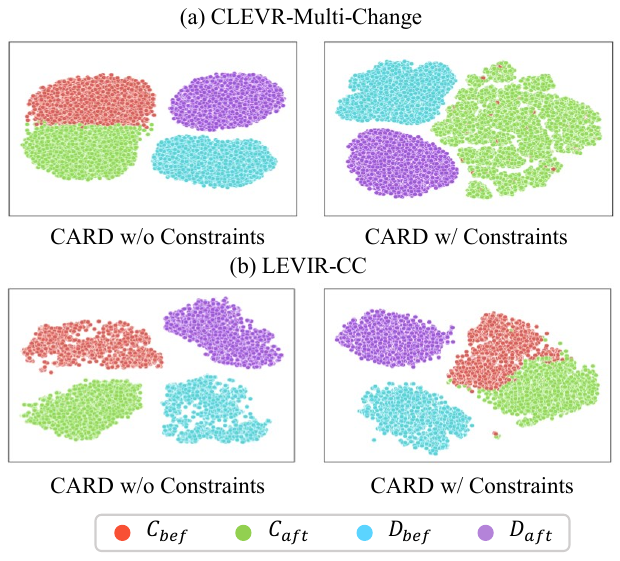} 
\caption{Visualization of context features on CLEVR-Multi-Change and LEVIR-CC. The red and green colors indicate common context features in ``before'' and ``after'' images, while blue and purple colors denote difference context features in  ``before'' and ``after'' images. }
\label{tsne}
\end{figure}

\begin{figure*}[hbtp]
\centering
\includegraphics[width=1\textwidth]{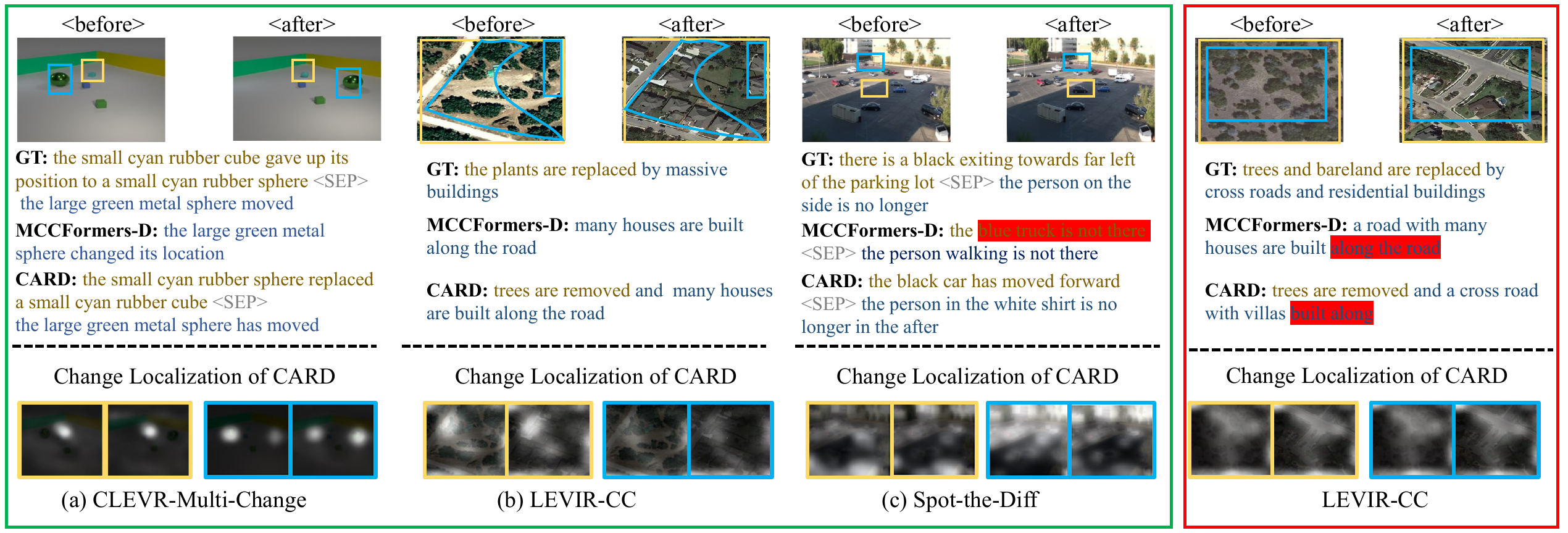} 
\caption{Qualitative examples on the three datasets. For each example, we visualize the captions generated by the SOTA method MCCFormers-D \cite{qiu2021describing} and our CARD, as well as the change localization of CARD. The successful cases of CARD are shown in the green box, while the sub-optimal case is shown in the red box. }
\label{3case}
\end{figure*}

\begin{figure}[hbtp]
\centering
\includegraphics[width=0.47\textwidth]{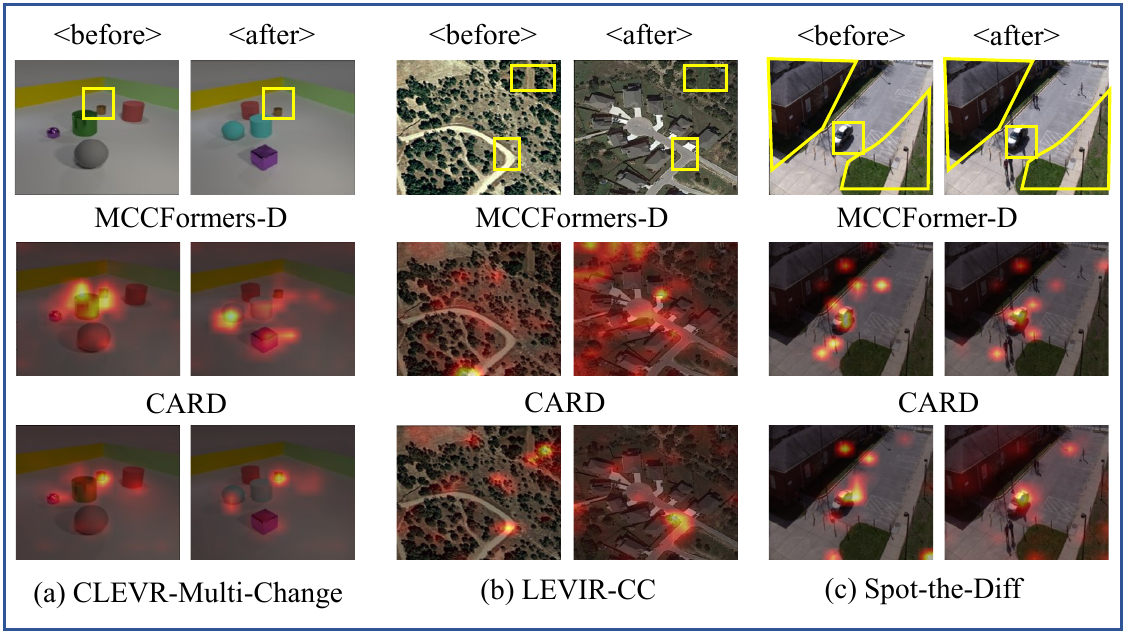} 
\caption{Visualization of alignment of common objects (shown in yellow boxes) on the three datasets, where the results are obtained by  MCCFormers-D and our CARD. }
\label{align}
\end{figure}
\subsubsection{Ablation Study for  Constrains} To study the effect of the  consistency constraint (CC) and independence constraint (IC), we make the ablation study  in Table \ref{constraint}. First, we enforce CC or IC respectively on CARD, and find that each of them improves the performance of CARD. Then,  we impose both constraints and observe that the results are enhanced significantly on two datasets. The increased results indicate that both constraint losses are essential to learn the context features. 

We further visualize the common/difference context features decoupled without/with the constraints, which are shown in Figure \ref{tsne}. Without constraints,  the common/difference context features cannot be well learned on two datasets.  With both constraints, the common context features are blended on CLEVR-Multi-Change, while the difference context features are learned better on LEVIR-CC. The results show that consistency constraint helps align distributions of shared properties, while independence constraint makes difference context features of two images more separable. 


\subsubsection{Generalization to unaligned image pairs}
To verify the generalization to unaligned image pairs under viewpoint changes, we compare CARD with two ablative variants on the CLEVR-Multi-Change dataset. (1) Direct Subtraction first performs direct subtraction between the features of two images to compute the locally difference features, which are then fed into a transformer decoder for multi-change captioning. (2) Feature Matching directly matches two image features to extract the locally common features and difference features for multi-change captioning.
The experimental results are shown in Table \ref{unaligned image}.

\begin{table}[htbp]
  \centering
  \small
  \caption{Verifying generalization to unaligned image pairs on CLEVR-Multi-Change.}
    \begin{tabular}{c|ccccc}
    \toprule
    Model & B & M & R & C & S \\
    \midrule
    Direct Subtraction & 53.3  & 42.5  & 56.3  & 350.0   & 74.6 \\
    Feature Matching & 54.7  & 43.6  & 56.7  & 362.3 & 75.6 \\
    CARD  & \textbf{56.7} & \textbf{45.2} & \textbf{57.4} & \textbf{391.6} & \textbf{76.9} \\
    \bottomrule
    \end{tabular}%
  \label{unaligned image}%
\end{table}%

It is noted that the performance of Feature Matching is better than that of Direct Subtraction on every metric, which validates the generalization of feature matching paradigm to the unaligned image pairs under viewpoint changes. Our proposed CARD outperforms both models by a large margin. This  not only indicates a better generalization of our method to unaligned image pairs, but also verifies the effectiveness of context-aware difference distilling to help capture genuine changes.

\subsection{Qualitative Analysis}

To obtain an overall evaluation of our method,  we conduct qualitative analysis on the three datasets.  The compared  method MCCFormers-D \cite{qiu2021describing} performs well on the three datasets, which directly correlates two images to predict locally common and  difference features.  In Figure \ref{align}, we visualize the alignment of common properties between two images, to validate whether context features help mine locally common information. We find that MCCFormers-D fails to align the common properties and even misjudges changed objects as unchanged objects. Instead, our CARD can better match the joint objects. For instance, in Figure \ref{align} (a), the unchanged object is only the brown object. MCCFormers-D wrongly identifies some changed objects as unchanged ones. By contrast, our CARD can pinpoint the unchanged brown object. This superiority benefits from the guidance of decoupled common context features, during the matching of two image features.

In Figure \ref{3case}, we further visualize the captions yielded by MCCFormers-D and CARD, as well as the change localization results from CARD. In the  first three cases,  MCCFormers-D either only describes one of the changes or misidentifies changed objects.  Contrarily, our CARD can accurately locate and describe all changed objects. Particularly, we notice that our method performs better in detecting subtle changes. For instance, in Figure \ref{3case} (c) that is from surveillance scene, the moved car and disappeared person are very tiny, and the car is occluded by the shadow of building. In this hard case, MCCFormers-D fails to recognize the moved car in the ``after’’ image, thus generating a wrong sentence. By contrary, our CARD can locate and describe this tiny change.
For the failure reason of MCCFormers-D, our conjecture is that it directly interacts two images, which cannot sufficiently identify locally unchanged features and compute the locally difference features. Besides, MCCFormers-D captures differences between two images only based on local features, which risks overlooking certain tiny changes with weak features. Compared with MCCFormers-D, the superiority of our method is mainly attributed to the guiding and augmenting of common and difference context features. Guided by the common context features, CARD models inter-image interaction to sufficiently mine locally common features and compute locally difference features. Further, the difference context features augment the locally difference features to ensure that all changes are distilled. Through this manner, the model can learn genuine changes for caption generation.
More qualitative examples  are shown in the appendix.


\section{Conclusion}
\label{conclusion}
In this paper, we  propose  the CARD to reason and describe genuine changes under various multi-change scenarios. CARD first decouples the common and difference context features from the image pair. Then, two kinds of  constraints are designed to ensure the alignment and discrepancy of the common and difference context features, respectively. Further, we use the common context features to guide the mining of locally common features  for deducing locally difference features. In addition, we  leverage the difference context features  to  augment the  locally difference features, thereby constructing an omni-representation of all  changes for multi-change captioning.
Extensive experiments conducted on the  three datasets  show that the proposed CARD outperforms the current state-of-the-art methods by a large margin. 

\section*{Limitations}
The last case in Figure \ref{3case} shows that the trees are replaced by a cross road with villas. Our CARD successfully locates the changed objects,  which validates the effectiveness of context-aware difference distilling. However, it yields a sub-optimal sentence that does not well express the change process: \textit{trees are removed and a cross road with villas built along}. A more proper sentence should be: \textit{trees are removed and a cross road with villas are built}. 
In the future work, we will try to introduce linguistic knowledge (\textit{e.g.}, syntactic dependencies between words) that regularizes the process of sentence generation, in order to generate the optimal sentences well elaborating the change process.

\section*{Acknowledgements}
This work was supported in part by National Key R\&D Program of China under Grant (2023YFB4502800), National Natural Science Foundation of China: 62322211, 61931008, 62236008, 62336008, U21B2038, 62225207, Fundamental Research Funds for the Central Universities (E2ET1104), ``Pionee'' and ``Leading Goose'' R\&D Program of Zhejiang Province (2024C01023, 2023C01030).

\bibliography{anthology,custom}

\begin{thebibliography}{43}
\expandafter\ifx\csname natexlab\endcsname\relax\def\natexlab#1{#1}\fi

\bibitem[{Anderson et~al.(2016)Anderson, Fernando, Johnson, and Gould}]{anderson2016spice}
Peter Anderson, Basura Fernando, Mark Johnson, and Stephen Gould. 2016.
\newblock Spice: Semantic propositional image caption evaluation.
\newblock In \emph{ECCV}, pages 382--398. Springer.

\bibitem[{Ba et~al.(2016)Ba, Kiros, and Hinton}]{ba2016layer}
Jimmy~Lei Ba, Jamie~Ryan Kiros, and Geoffrey~E Hinton. 2016.
\newblock Layer normalization.
\newblock \emph{arXiv preprint arXiv:1607.06450}.

\bibitem[{Banerjee and Lavie(2005)}]{banerjee2005meteor}
Satanjeev Banerjee and Alon Lavie. 2005.
\newblock Meteor: An automatic metric for mt evaluation with improved correlation with human judgments.
\newblock In \emph{Proceedings of the acl workshop on intrinsic and extrinsic evaluation measures for machine translation and/or summarization}, pages 65--72.

\bibitem[{Chang and Ghamisi(2023)}]{Change2023changes}
Shizhen Chang and Pedram Ghamisi. 2023.
\newblock Changes to captions: An attentive network for remote sensing change captioning.
\newblock \emph{IEEE Transactions on Image Processing}, 32:6047--6060.

\bibitem[{Chen et~al.(2015)Chen, Fang, Lin, Vedantam, Gupta, Doll{\'a}r, and Zitnick}]{chen2015microsoft}
Xinlei Chen, Hao Fang, Tsung-Yi Lin, Ramakrishna Vedantam, Saurabh Gupta, Piotr Doll{\'a}r, and C~Lawrence Zitnick. 2015.
\newblock Microsoft coco captions: Data collection and evaluation server.
\newblock \emph{arXiv preprint arXiv:1504.00325}.

\bibitem[{Cong et~al.(2022)Cong, Li, Liu, Tu, Qin, Zhang, Yan, Wang, and Jiang}]{cong2022ls}
Gaoxiang Cong, Liang Li, Zhenhuan Liu, Yunbin Tu, Weijun Qin, Shenyuan Zhang, Chengang Yan, Wenyu Wang, and Bin Jiang. 2022.
\newblock Ls-gan: iterative language-based image manipulation via long and short term consistency reasoning.
\newblock In \emph{ACM MM}, pages 4496--4504.

\bibitem[{Cong et~al.(2023)Cong, Li, Qi, Zha, Wu, Wang, Jiang, Yang, and Huang}]{cong2023learning}
Gaoxiang Cong, Liang Li, Yuankai Qi, Zheng-Jun Zha, Qi~Wu, Wenyu Wang, Bin Jiang, Ming-Hsuan Yang, and Qingming Huang. 2023.
\newblock Learning to dub movies via hierarchical prosody models.
\newblock In \emph{CVPR}, pages 14687--14697.

\bibitem[{Guo et~al.(2022)Guo, Wang, and Laaksonen}]{guo2022clip4idc}
Zixin Guo, Tzu-Jui Wang, and Jorma Laaksonen. 2022.
\newblock Clip4idc: Clip for image difference captioning.
\newblock In \emph{AACL}, pages 33--42.

\bibitem[{He et~al.(2016)He, Zhang, Ren, and Sun}]{he2016deep}
Kaiming He, Xiangyu Zhang, Shaoqing Ren, and Jian Sun. 2016.
\newblock Deep residual learning for image recognition.
\newblock In \emph{CVPR}, pages 770--778.

\bibitem[{Hosseinzadeh and Wang(2021)}]{hosseinzadeh2021image}
Mehrdad Hosseinzadeh and Yang Wang. 2021.
\newblock Image change captioning by learning from an auxiliary task.
\newblock In \emph{CVPR}, pages 2725--2734.

\bibitem[{Hoxha et~al.(2022)Hoxha, Chouaf, Melgani, and Smara}]{hoxha2022change}
Genc Hoxha, Seloua Chouaf, Farid Melgani, and Youcef Smara. 2022.
\newblock Change captioning: A new paradigm for multitemporal remote sensing image analysis.
\newblock \emph{IEEE Transactions on Geoscience and Remote Sensing}, 60:1--14.

\bibitem[{Huang et~al.(2022)Huang, Liang, Wei, Cai, Liang, Leung, and Li}]{huang2022image}
Qingbao Huang, Yu~Liang, Jielong Wei, Yi~Cai, Hanyu Liang, Ho-fung Leung, and Qing Li. 2022.
\newblock Image difference captioning with instance-level fine-grained feature representation.
\newblock \emph{IEEE Transactions on Multimedia}, 24:2004--2017.

\bibitem[{Jhamtani and Berg-Kirkpatrick(2018)}]{jhamtani2018learning}
Harsh Jhamtani and Taylor Berg-Kirkpatrick. 2018.
\newblock Learning to describe differences between pairs of similar images.
\newblock In \emph{EMNLP}, pages 4024--4034.

\bibitem[{Kim et~al.(2021)Kim, Kim, Lee, Park, and Kim}]{kim2021agnostic}
Hoeseong Kim, Jongseok Kim, Hyungseok Lee, Hyunsung Park, and Gunhee Kim. 2021.
\newblock Agnostic change captioning with cycle consistency.
\newblock In \emph{ICCV}, pages 2095--2104.

\bibitem[{Kingma and Ba(2014)}]{kingma2014adam}
Diederik~P Kingma and Jimmy Ba. 2014.
\newblock Adam: A method for stochastic optimization.
\newblock \emph{arXiv preprint arXiv:1412.6980}.

\bibitem[{Lin(2004)}]{lin2004rouge}
Chin-Yew Lin. 2004.
\newblock Rouge: A package for automatic evaluation of summaries.
\newblock In \emph{Text summarization branches out}, pages 74--81.

\bibitem[{Liu et~al.(2023{\natexlab{a}})Liu, Yang, Qi, Zou, and Shi}]{liu2023progressive}
Chenyang Liu, Jiajun Yang, Zipeng Qi, Zhengxia Zou, and Zhenwei Shi. 2023{\natexlab{a}}.
\newblock Progressive scale-aware network for remote sensing image change captioning.
\newblock \emph{IGARSS}.

\bibitem[{Liu et~al.(2022)Liu, Zhao, Chen, Zou, and Shi}]{liu2022remote}
Chenyang Liu, Rui Zhao, Hao Chen, Zhengxia Zou, and Zhenwei Shi. 2022.
\newblock Remote sensing image change captioning with dual-branch transformers: A new method and a large scale dataset.
\newblock \emph{IEEE Transactions on Geoscience and Remote Sensing}, 60:1--20.

\bibitem[{Liu et~al.(2023{\natexlab{b}})Liu, Zhao, Chen, Qi, Zou, and Shi}]{liu2023Decoupling}
Chenyang Liu, Rui Zhao, Jianqi Chen, Zipeng Qi, Zhengxia Zou, and Zhenwei Shi. 2023{\natexlab{b}}.
\newblock A decoupling paradigm with prompt learning for remote sensing image change captioning.
\newblock \emph{IEEE Transactions on Geoscience and Remote Sensing}, 61:1--18.

\bibitem[{Liu et~al.(2021)Liu, Yin, Wu, Ge, Zhang, and Sun}]{liu2021contrastive}
Fenglin Liu, Changchang Yin, Xian Wu, Shen Ge, Ping Zhang, and Xu~Sun. 2021.
\newblock Contrastive attention for automatic chest x-ray report generation.
\newblock In \emph{Findings of ACL}, pages 269--280.

\bibitem[{Oord et~al.(2018)Oord, Li, and Vinyals}]{oord2018representation}
Aaron van~den Oord, Yazhe Li, and Oriol Vinyals. 2018.
\newblock Representation learning with contrastive predictive coding.
\newblock \emph{arXiv preprint arXiv:1807.03748}.

\bibitem[{Papineni et~al.(2002)Papineni, Roukos, Ward, and Zhu}]{papineni2002bleu}
Kishore Papineni, Salim Roukos, Todd Ward, and Wei-Jing Zhu. 2002.
\newblock Bleu: a method for automatic evaluation of machine translation.
\newblock In \emph{ACL}, pages 311--318.

\bibitem[{Park et~al.(2019)Park, Darrell, and Rohrbach}]{park2019robust}
Dong~Huk Park, Trevor Darrell, and Anna Rohrbach. 2019.
\newblock Robust change captioning.
\newblock In \emph{ICCV}, pages 4624--4633.

\bibitem[{Paszke et~al.(2019)Paszke, Gross, Massa, Lerer, Bradbury, Chanan, Killeen, Lin, Gimelshein, Antiga et~al.}]{paszke2019pytorch}
Adam Paszke, Sam Gross, Francisco Massa, Adam Lerer, James Bradbury, Gregory Chanan, Trevor Killeen, Zeming Lin, Natalia Gimelshein, Luca Antiga, et~al. 2019.
\newblock Pytorch: An imperative style, high-performance deep learning library.
\newblock In \emph{NeurIPS}, pages 8024--8035.

\bibitem[{Qiu et~al.(2023)Qiu, Sun, Matsuzawa, Iwata, and Kataoka}]{qiu2023graph}
Yue Qiu, Yanjun Sun, Fumiya Matsuzawa, Kenji Iwata, and Hirokatsu Kataoka. 2023.
\newblock Graph representation for order-aware visual transformation.
\newblock In \emph{CVPR}, pages 22793--22802.

\bibitem[{Qiu et~al.(2021)Qiu, Yamamoto, Nakashima, Suzuki, Iwata, Kataoka, and Satoh}]{qiu2021describing}
Yue Qiu, Shintaro Yamamoto, Kodai Nakashima, Ryota Suzuki, Kenji Iwata, Hirokatsu Kataoka, and Yutaka Satoh. 2021.
\newblock Describing and localizing multiple changes with transformers.
\newblock In \emph{ICCV}, pages 1971--1980.

\bibitem[{Rotstein et~al.(2024)Rotstein, Bensa{\"\i}d, Brody, Ganz, and Kimmel}]{rotstein2024fusecap}
Noam Rotstein, David Bensa{\"\i}d, Shaked Brody, Roy Ganz, and Ron Kimmel. 2024.
\newblock Fusecap: Leveraging large language models for enriched fused image captions.
\newblock In \emph{WACV}, pages 5689--5700.

\bibitem[{Shi et~al.(2020)Shi, Yang, Gu, Joty, and Cai}]{shi2020finding}
Xiangxi Shi, Xu~Yang, Jiuxiang Gu, Shafiq Joty, and Jianfei Cai. 2020.
\newblock Finding it at another side: A viewpoint-adapted matching encoder for change captioning.
\newblock In \emph{ECCV}, pages 574--590.

\bibitem[{Song et~al.(2007)Song, Smola, Gretton, Borgwardt, and Bedo}]{song2007supervised}
Le~Song, Alex Smola, Arthur Gretton, Karsten~M Borgwardt, and Justin Bedo. 2007.
\newblock Supervised feature selection via dependence estimation.
\newblock In \emph{ICML}, pages 823--830.

\bibitem[{Tu et~al.(2023{\natexlab{a}})Tu, Li, Su, Du, Lu, and Huang}]{tu2023viewpoint}
Yunbin Tu, Liang Li, Li~Su, Junping Du, Ke~Lu, and Qingming Huang. 2023{\natexlab{a}}.
\newblock Viewpoint-adaptive representation disentanglement network for change captioning.
\newblock \emph{IEEE Transactions on Image Processing}, 32:2620--2635.

\bibitem[{Tu et~al.(2022)Tu, Li, Su, Gao, Yan, Zha, Yu, and Huang}]{tu20222}
Yunbin Tu, Liang Li, Li~Su, Shengxiang Gao, Chenggang Yan, Zheng-Jun Zha, Zhengtao Yu, and Qingming Huang. 2022.
\newblock I2 transformer: Intra-and inter-relation embedding transformer for tv show captioning.
\newblock \emph{IEEE Transactions on Image Processing}, 31:3565--3577.

\bibitem[{Tu et~al.(2023{\natexlab{b}})Tu, Li, Su, Lu, and Huang}]{tu2023neighborhood}
Yunbin Tu, Liang Li, Li~Su, Ke~Lu, and Qingming Huang. 2023{\natexlab{b}}.
\newblock Neighborhood contrastive transformer for change captioning.
\newblock \emph{IEEE Transactions on Multimedia}, pages 1--12.

\bibitem[{Tu et~al.(2024)Tu, Li, Su, Zha, and Huang}]{tu2024smart}
Yunbin Tu, Liang Li, Li~Su, Zheng-Jun Zha, and Qingming Huang. 2024.
\newblock Smart: Syntax-calibrated multi-aspect relation transformer for change captioning.
\newblock \emph{IEEE Transactions on Pattern Analysis and Machine Intelligence}.

\bibitem[{Tu et~al.(2023{\natexlab{c}})Tu, Li, Su, Zha, Yan, and Huang}]{tu2023self}
Yunbin Tu, Liang Li, Li~Su, Zheng-Jun Zha, Chenggang Yan, and Qingming Huang. 2023{\natexlab{c}}.
\newblock Self-supervised cross-view representation reconstruction for change captioning.
\newblock In \emph{ICCV}, pages 2805--2815.

\bibitem[{Tu et~al.(2021{\natexlab{a}})Tu, Li, Yan, Gao, and Yu}]{tu-etal-2021-r}
Yunbin Tu, Liang Li, Chenggang Yan, Shengxiang Gao, and Zhengtao Yu. 2021{\natexlab{a}}.
\newblock {R}{\^{}}3{N}et:relation-embedded representation reconstruction network for change captioning.
\newblock In \emph{EMNLP}, pages 9319--9329.

\bibitem[{Tu et~al.(2021{\natexlab{b}})Tu, Yao, Li, Lou, Gao, Yu, and Yan}]{tu2021semantic}
Yunbin Tu, Tingting Yao, Liang Li, Jiedong Lou, Shengxiang Gao, Zhengtao Yu, and Chenggang Yan. 2021{\natexlab{b}}.
\newblock Semantic relation-aware difference representation learning for change captioning.
\newblock In \emph{Findings of ACL}, pages 63--73.

\bibitem[{Vaswani et~al.(2017)Vaswani, Shazeer, Parmar, Uszkoreit, Jones, Gomez, Kaiser, and Polosukhin}]{vaswani2017attention}
Ashish Vaswani, Noam Shazeer, Niki Parmar, Jakob Uszkoreit, Llion Jones, Aidan~N Gomez, {\L}ukasz Kaiser, and Illia Polosukhin. 2017.
\newblock Attention is all you need.
\newblock In \emph{NeurIPS}, pages 5998--6008.

\bibitem[{Vedantam et~al.(2015)Vedantam, Lawrence~Zitnick, and Parikh}]{vedantam2015cider}
Ramakrishna Vedantam, C~Lawrence~Zitnick, and Devi Parikh. 2015.
\newblock Cider: Consensus-based image description evaluation.
\newblock In \emph{CVPR}, pages 4566--4575.

\bibitem[{Yang et~al.(2023)Yang, Peng, Wang, Xu, Ye, Li, Huang, Huang, Li, and Zhang}]{yang2023transforming}
Xu~Yang, Jiawei Peng, Zihua Wang, Haiyang Xu, Qinghao Ye, Chenliang Li, Songfang Huang, Fei Huang, Zhangzikang Li, and Yu~Zhang. 2023.
\newblock Transforming visual scene graphs to image captions.
\newblock In \emph{ACL}, pages 12427--12440.

\bibitem[{Yao et~al.(2022)Yao, Wang, and Jin}]{yao2022image}
Linli Yao, Weiying Wang, and Qin Jin. 2022.
\newblock Image difference captioning with pre-training and contrastive learning.
\newblock In \emph{AAAI}.

\bibitem[{Yue et~al.(2024)Yue, Tu, Li, Gao, and Yu}]{yue2024multi}
Shengbin Yue, Yunbin Tu, Liang Li, Shengxiang Gao, and Zhengtao Yu. 2024.
\newblock Multi-grained representation aggregating transformer with gating cycle for change captioning.
\newblock \emph{ACM Transactions on Multimedia Computing, Communications and Applications}.

\bibitem[{Yue et~al.(2023)Yue, Tu, Li, Yang, Gao, and Yu}]{yue2023i3n}
Shengbin Yue, Yunbin Tu, Liang Li, Ying Yang, Shengxiang Gao, and Zhengtao Yu. 2023.
\newblock I3n: Intra- and inter-representation interaction network for change captioning.
\newblock \emph{IEEE Transactions on Multimedia}, pages 1--14.

\bibitem[{Zhao and Xiong(2024)}]{zhao2024cooperative}
Kai Zhao and Wei Xiong. 2024.
\newblock Cooperative connection transformer for remote sensing image captioning.
\newblock \emph{IEEE Transactions on Geoscience and Remote Sensing}.

\end{thebibliography}
\bibliographystyle{acl_natbib}

\appendix

\section{Appendix}
\label{sec:appendix}

In this appendix, we provide more details about the caption generation and more experimental results, as well as the qualitative analyses.
\subsection{Caption Generation}
\label{caption generation supp}
After obtaining the representation of all changes $ \tilde{X}_{d} \in \mathbb{R}^{hw \times d}$, we use a transformer decoder to decode it into target sentences. First, we obtain the embedding features of all $m$ words of these sentence $E[S]\in \mathbb{R}^{m \times d}$, where each sentence is separated by a special token [SEP]. Then, we use the masked self-attention \cite{vaswani2017attention} to model  relationships among these word features, which  is defined as:
\begin{equation}
 {\hat E[S]}=\text { LN }(E[S]+\text { MHSA }(E[S],E[S],E[S])),
\end{equation}
where LN is short for layer normalization \cite{ba2016layer}.
Next, we model the interaction between these word features and difference representation $\tilde{X}_{d}$ by multi-head cross-attention (MHCA) \cite{vaswani2017attention}, so as to locate the most related difference features $\tilde H$:
\begin{equation}
\tilde H=\text { LN } ({E[\hat S]}+\text { MHCA }({E[\hat S]},\tilde{X}_{d},\tilde{X}_{d})).
\end{equation}
Subsequently, we feed the selected features $\tilde H$ into a feed-forward network to obtain the enhanced difference representation, denoted as $\hat H \in \mathbb{R}^{m\times d}$.
\begin{equation}
\hat H=\text { LN}((\tilde H+\operatorname {FFN}(\tilde H)).
\end{equation}
Finally, the probability distributions of   words in these sentences are calculated via a single hidden layer:
\begin{equation}
S=\operatorname{Softmax}\left(\hat H W_{s}+{b}_{s}\right),
\end{equation}
where $W_{s}\in \mathbb{R}^{d \times u}$ and $b_{s} \in \mathbb{R}^{u}$ are the learnable parameters. $u$ is the dimension of vocabulary size.

\subsection{Experiments}
\subsubsection{Implementation Details}
For fair-comparison, we follow the previous multi-change captioning methods \cite{jhamtani2018learning,qiu2021describing,liu2022remote} to use a pre-trained ResNet-101  \citep{he2016deep} to extract the local features of a pair of images, with the dimension of 1024 $\times$ 14 $\times$ 14. We project them into a lower dimension of 512, while the dimension of trainable [CLS] features is also set to 512. The hidden size of overall model and word embedding size are set to 512 and 300, respectively. Temperature $\tau$ in Eq. (2) of main paper is set to 0.07. The attention layers in CARD is set to 1 on the CLEVR-Multi-Change and Spot-the-Diff datasets; and 3 on the LEVIR-CC dataset.
We train the model with PyTorch \cite{paszke2019pytorch} on a single RTX 3090 GPU, and use Adam optimizer \citep{kingma2014adam} to minimize the negative log-likelihood loss of Eq. (\ref{cross-entropy}) in the main paper. The training details about batch size and learning rate are shown in Table \ref{batch}. The used training resources about time and GPU memory are shown in Table \ref{time}. We find that training CARD does not require much more time and GPU memory. Hence, it would be easily re-implemented by other researchers  and be a strong baseline for the future works.

\begin{table}[h]
\small
\centering
	\caption{The training details of CARD on the three datasets.}	
    \begin{tabular}{c|c|c}
   \hline
          & batch size & learning rate \\
    \hline
    CLEVR-Multi-Change & 128  & 2 $\times$ $10^{-4}$ \\
    LEVIR-CC & 64 &  1 $\times$ $10^{-4}$ \\
    Spot-the-Diff & 32 &  2 $\times$ $10^{-4}$ \\
    \hline
    \end{tabular}%
\label{batch}%
	\end{table} 

 \begin{table}[h]
			\centering
			\small
     \caption{The used training resources of CARD on the three datasets.}
    \begin{tabular}{c|c|c}
    \hline
          & Training Time & GPU Memory \\
    \hline
    CLEVR-Multi-Change & 71 minutes  & 9.2 GB  \\
    LEVIR-CC & 120 minutes  & 8 GB \\
    Spot-the-Diff & 20 minutes  & 3.9 GB \\
    \hline
    \end{tabular}%
  
  \label{time}%
		\end{table}

\begin{table}[htbp]
  \centering
  \small
  \caption{Effects of  $\lambda_c$ on  CLEVR-Multi-Change.}
    \begin{tabular}{c|c|c|c|c|c|c}
    \toprule
    Model &   $\lambda_c$    & B     & M     & R  & S   & \cellcolor[rgb]{ .859,  .859,  .859}C  \\
    \midrule
    CARD  & 0     & 56.2  & 45.3  & 57.3 & 76.9 & \cellcolor[rgb]{ .859,  .859,  .859}387.8  \\
    CARD  & 0.1   & 56.6  & 45.2  & \textbf{57.4} & \textbf{77.0} & \cellcolor[rgb]{ .859,  .859,  .859}391.2  \\
    CARD  & 0.2   & 56.6 & \textbf{45.3} & 57.3 & \textbf{77.0} & \cellcolor[rgb]{ .859,  .859,  .859}390.7  \\
    CARD  & 0.3   & \textbf{56.7} & 45.2 & \textbf{57.4} & 76.9 & \cellcolor[rgb]{ .859,  .859,  .859}\textbf{391.6}  \\
    CARD  & 0.4   & 56.5  & 45.2  & 57.0 & 76.8 & \cellcolor[rgb]{ .859,  .859,  .859}388.8  \\
    CARD  & 0.5   & 56.6 & \textbf{45.3} &\textbf{ 57.4} & 76.9 & \cellcolor[rgb]{ .859,  .859,  .859}390.7  \\
    \bottomrule
    \end{tabular}%
    
  \label{trade}%
\end{table}%

\begin{table}[htbp]
  \centering
  \small
   \caption{Effects of  $\lambda_c$ on the  LEVIR-CC dataset.}
    \begin{tabular}{c|c|c|c|c|c}
    \toprule
    Model &   $\lambda_c$    & B     & M     & R    & \cellcolor[rgb]{ .859,  .859,  .859}C  \\
    \midrule
    CARD  & 0     & 55.9  & 35.6  & 72.3  & \cellcolor[rgb]{ .859,  .859,  .859}132.2  \\
    CARD  & 0.1   & \textbf{65.4} & \textbf{40.0} & \textbf{74.6} & \cellcolor[rgb]{ .859,  .859,  .859}\textbf{137.9}  \\
    CARD  & 0.2   & 63.7 & 39.3 & 73.3  & \cellcolor[rgb]{ .859,  .859,  .859}132.9  \\
    CARD  & 0.3   & 54.8  & 34.4 & 72.7   & \cellcolor[rgb]{ .859,  .859,  .859} 137.5  \\
    CARD  & 0.4   & 63.2  & 38.3  & 73.6  & \cellcolor[rgb]{ .859,  .859,  .859}137.5  \\
    CARD  & 0.5   & 59.5 & 36.9 & 72.9 & \cellcolor[rgb]{ .859,  .859,  .859}135.1  \\
    \bottomrule
    \end{tabular}%
  \label{trade_levir}%
\end{table}%

\begin{table}[htbp]
  \centering
  \small
  \caption{Effects of  $\lambda_c$ on the Spot-the-Diff dataset.}
    \begin{tabular}{c|c|c|c|c|c|c}
    \toprule
    Model &   $\lambda_c$    & B     & M     & R  & S   & \cellcolor[rgb]{ .859,  .859,  .859}C  \\
    \midrule
    CARD  & 0     & 4.3  & 10.6  & 23.3 & 13.0 & \cellcolor[rgb]{ .859,  .859,  .859}15.8  \\
    CARD  & 0.001  & \textbf{6.6} & 10.8  & \textbf{26.9} & \textbf{17.8} & \cellcolor[rgb]{ .859,  .859,  .859}\textbf{32.4}   \\
    CARD  & 0.002   & 4.3 & 9.8 & 23.9 & 15.1 & \cellcolor[rgb]{ .859,  .859,  .859}23.0  \\
    CARD  & 0.003   & 6.2  & 9.5  & 25.7  & 15.7 & \cellcolor[rgb]{ .859,  .859,  .859}28.4  \\
    CARD  & 0.004   & 5.0  & 11.0 & 24.4 & 15.2 & \cellcolor[rgb]{ .859,  .859,  .859}20.5  \\
    CARD  & 0.005   & 5.1 & \textbf{11.4} & 24.0 & 15.6 & \cellcolor[rgb]{ .859,  .859,  .859}18.3  \\
    \bottomrule
    \end{tabular}%
    
  \label{trade_spot}%
\end{table}%

\subsection{Study on the Trade-off Parameter $\lambda_c$}
We discuss the influence of trade-off parameter $\lambda_c$ in Eq. (\ref{cross-entropy}). The results under varied values are shown in Table \ref{trade}-\ref{trade_spot}. Note
that  $\lambda_c$=0 means there are no constraints upon the decoupled results.  We observe that imposing the constraint losses does improve the model's performance. Besides, there is a trade-off between the captioning loss and constraint losses, because too large $\lambda_c$ may lead to deterioration in caption performance. Based on the results, we choose the value as 0.3 on CLEVR-Multi-Change, 0.1 and 0.001 on the LEVIR-CC and Spot-the-Diff, respectively.


\begin{figure*}[t]
\centering
\includegraphics[width=1\textwidth]{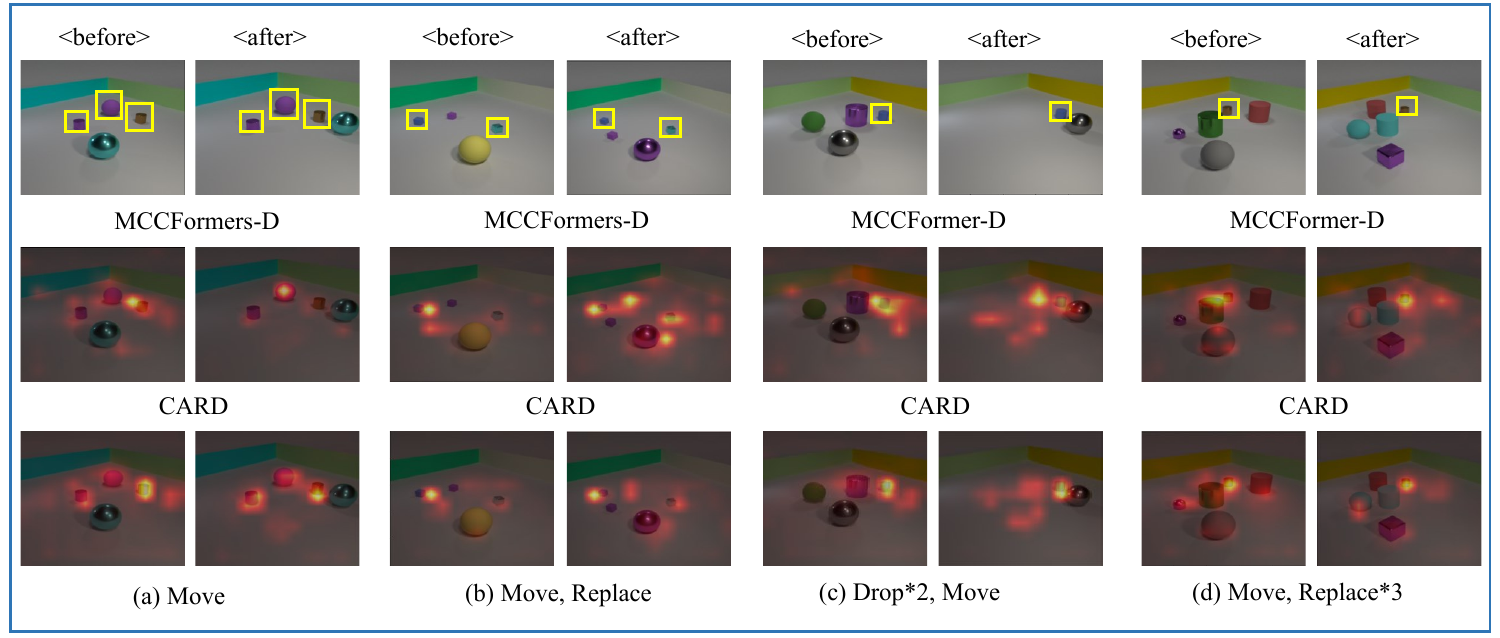} 
\caption{Visualization of common objects matching  on the CLEVR-Multi-Change dataset under one-to-four changes. For each example, we  visualize the matching results by the state-of-the-art method MCCFormers-D \cite{qiu2021describing} and our CARD.  The common objects are shown in the yellow boxes.     }
\label{change_align}
\end{figure*}

\begin{figure*}[t]
\centering
\includegraphics[width=1\textwidth]{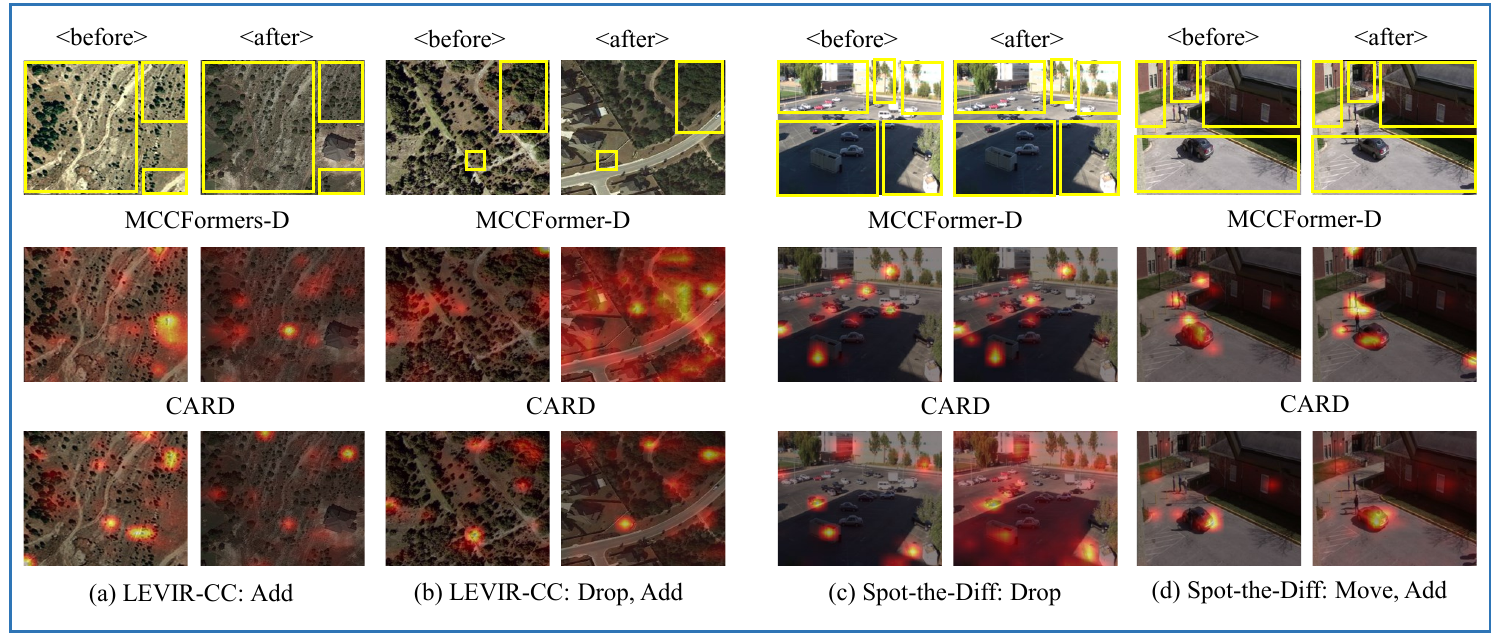} 
\caption{Visualization of common objects matching on the LEVIR-CC and Spot-the-Diff datasets under varied changes. For each example, we  visualize the matching results by the state-of-the-art method MCCFormers-D \cite{qiu2021describing} and our CARD.  The common objects are shown in the yellow boxes.   }
\label{spot_align}
\end{figure*}

\begin{figure*}[t]
\centering
\includegraphics[width=1\textwidth]{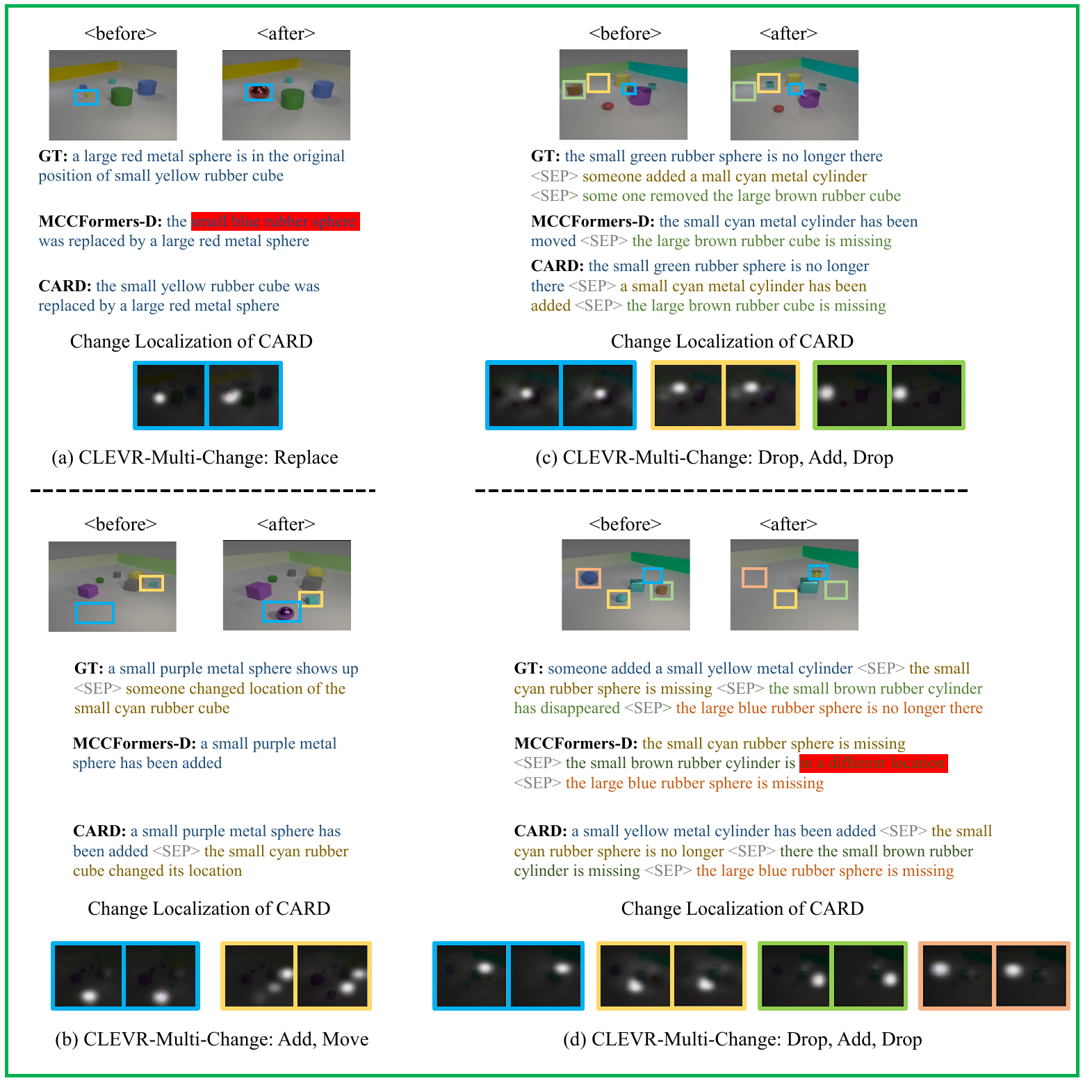} 
\caption{Qualitative examples on the CLEVR-Multi-Change dataset under one-to-four changes. For each example, we visualize the captions generated by the state-of-the-art method MCCFormers-D \cite{qiu2021describing} and our CARD, as well as the change localization of CARD. The changed objects are shown in the colored boxes. }
\label{change_text}
\end{figure*}

\begin{figure*}[t]
\centering
\includegraphics[width=1\textwidth]{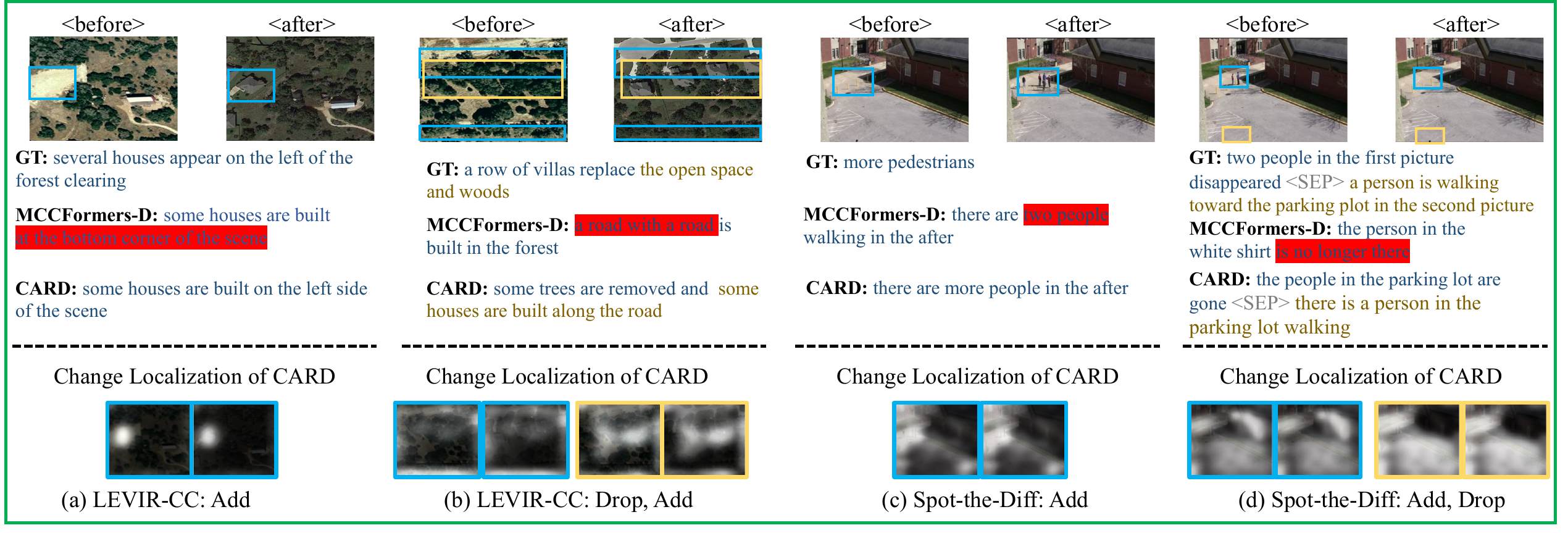} 
\caption{Qualitative examples on the LEVIR-CC and Spot-the-Diff datasets under varied changes. For each example, we visualize the captions generated by the state-of-the-art method MCCFormers-D \cite{qiu2021describing} and our CARD, as well as the change localization of CARD. The changed objects are shown in the colored boxes. }
\label{spot_text}
\end{figure*}

\subsection{Qualitative Analysis}
In this appendix, we will show more qualitative examples on the CLEVR-Multi-Change, LEVIR-CC, and Spot-the-Diff datasets, which are shown in Figure \ref{change_align}-\ref{spot_text}. 
To intuitively understand whether the common context features help mine reliable common properties, we visualize the alignment of common properties between two images on the three datasets, which are shown in Figure \ref{change_align}-\ref{spot_align}. The compared method is  MCCFormers-D \cite{qiu2021describing}, which has the stable performance on the three datasets.  From these examples, we can observe that MCCFormers-D is unable to align the common properties properly and even misjudges changed objects as unchanged objects. Compared with it, the proposed CARD can better match the common objects, so as to validate the effectiveness of the decoupled common context features. 
 
 Further, in Figure \ref{change_text}-\ref{spot_text}, we  visualize the captions yielded by MCCFormers-D and CARD, as well as the change localization results from CARD on the three datasets. In Figure \ref{change_text}, on the CLEVR-Multi-Change dataset under one-to-four changes, we find that  MCCFormers-D either only describes partial changes or misidentifies changed objects. Instead, the proposed CARD is able to accurately locate and describe all changed objects. In Figure \ref{spot_text}, on the LEVIR-CC and Spot-the-Diff datasets, it is noted that MCCFormers-D fails to describe all the changes within each image pair that is from the real-word environment.  Different it, our CARD is capable of  locating and describing all changed objects, which show a good generalization and robustness of our method. The superiority benefits from that the proposed CARD can provide the model with a overview of potential changed/unchanged semantics within an image pair, which helps learn genuine changes for caption generation.

\end{document}